\documentclass{llncs}
\usepackage{llncsdoc}
\usepackage{graphicx,epsfig,subfig}
\usepackage{url}

\usepackage{fancyhdr}
\begin{document}
\title{VSCAN: An Enhanced Video Summarization using Density-based Spatial Clustering}
\author{Karim M. Mohamed \and Mohamed A. Ismail \and Nagia M. Ghanem }
\institute{Computer and Systems Engineering Department\\Faculty of Engineering, Alexandria University\\ Alexandria, Egypt}
\maketitle

\begin{abstract}
In this paper, we present VSCAN, a novel approach for generating static video summaries. This approach is based on a modified DBSCAN clustering algorithm to summarize the video content utilizing both color and texture features of the video frames. The paper also introduces an enhanced evaluation method that depends on color and texture features. Video Summaries generated by VSCAN are compared with summaries generated by other approaches found in the literature and those created by users. Experimental results indicate that the video summaries generated by VSCAN have a higher quality than those generated by other approaches.


\end{abstract}

\section{Introduction }
The revolution in digital video has been driven by the rapid development of computer infrastructure in various areas such as improved processing power, enhanced and cheaper capacity of storage devices, and faster networks. This revolution has brought many new applications and as a consequence research into new technologies that aim at improving the effectiveness and efficiency of video acquisition, archiving, cataloguing and indexing as well as increasing the usability of stored videos. This leads to the requirement of efficient management of video data such as video summarization. 

A video summary is defined as a sequence of still pictures that represent the content of a video in such a way that the respective target group is rapidly provided with concise information about the content, while the essential message of the original video is preserved \cite{pfeiffer1996abstracting}.

Over the past years, various approaches and techniques have been proposed towards the summarization of video content. However there are many drawbacks for these approaches. First, most of video summarization approaches that achieved a relatively high quality are based only on a single visual descriptor such as the color of the video frames; while other descriptors like texture is not considered. Second, clustering algorithms used in current video summarization techniques could not detect noise frames automatically; instead some of these techniques have to detect noise frames using separate methods which require additional computation. Third, the current video summarization approaches depend on special input parameters that may not be suitable for all cases. For example, many approaches utilizes k-means partitioning-based clustering algorithm that requires the number of clusters as an input; while the number of clusters is not related to the perceptual content of the automatic video summary. To overcome this problem, additional stage is required to filter key frames which increases the complexity of the video summarization process and makes using the clustering algorithm inefficient. Finally, current evaluation methods depend only on color features for comparing different summaries and do not consider other features like texture; which gives a less perceptual assessment of the quality of video summaries.

In this paper, we present VSCAN, an enhanced approach for generating static video summaries that operates on the whole video clip. It relies on clustering color and texture features extracted from the video frames using a modified DBSCAN \cite{ester1996density} algorithm to summarize the video content which overcomes the drawbacks of the other approaches. Also, we introduce an enhanced evaluation method that depends on color and texture features. VSCAN approach is evaluated using the enhanced evaluation method and the experimental results show that VSCAN produces video summaries with higher quality than those generated by other approaches.

The rest of this paper is organized as follows. Section 2 introduces some related work. Section 3 presents VSCAN approach and shows how to apply it to summarize a video sequence. Section 4 illustrates the evaluation method and reports the results of our experiments. Finally, we offer our conclusions and directions for future work in Section 5. 

\section{Related Work}		
A comprehensive review of video summarization approaches can be found in \cite{truong2007video}. Some of the main approaches and techniques related to static video summarization which can be found in the literature are briefly discussed next. 

In \cite{mundur2006keyframe}, an approach based on clustering the video frames using the Delaunay Triangulation (DT) is developed. The first step in this apporach is pre-sampling the frames of the input video. Then, the video frames are represented by a color histogram in the HSV color space and the Principal Component Analysis (PCA) is applied on the color feature matrix to reduce its dimensionality. After that, the Delaunay diagram is built and clusters are formed by separating edges in the Delaunay diagram. Finally, for each cluster, the frame that is closest to its center is selected as the key frame.  

In \cite{furini2010stimo}, an approach called STIMO (STIll and MOving Video Storyboard) is introduced. This approach is designed to produce on-the-fly video storyboards and it is composed of three phases. In the first phase, the video frames are pre-sampled and then feature vectors are extracted from the selected video frames by computing a color histogram in the HSV color space. In the second phase, a clustering method based on the Furthest-Point-First (FPF) algorithm is applied. To estimate the number of clusters, the pairwise distance of consecutive frames is computed using Generalized Jaccard Distance (GJD). Finally, a post-processing step is performed for removing noise video frames. 

In \cite{de2011vsumm}, an approach called VSUMM (Video SUMMarization) is presented. In the first step, the video frames are pre-sampled by selecting one frame per second. In the second step, the color features of video frames are extracted from Hue component only in the HSV color space. In the third step, the meaningless frames are eliminated. In the fourth step, the frames are clustered using k-means algorithm where the number of clusters is estimated by computing the pairwise Euclidean distances between video frames and a key frame is extracted from each cluster. Finally, another extra step occurs in which the key frames are compared among themselves through color histogram to eliminate those similar key frames in the produced summaries.

\section{VSCAN Approach}
\figurename{1} shows the steps of VSCAN approach to produce static video summaries. First, the original video is pre-sampled (step 1). Second, color features are extracted using color histogram in HSV color space (step 2). Third, texture features are extracted using a two-dimensional Haar wavelet transform in HSV color space (step 3). In step 4, video frames are clustered by a modified DBSCAN clustering algorithm. Then, in step 5, the key frames are selected. Finally, the extracted key frames are arranged in the original order of appearance in the video to facilitate the visual understanding of the result. These steps are explained in more details in the following subsections.

\subsection{Video Frames Pre-sampling}
The first step towards video summarization is pre-sampling the original video which aims to reduce the number of frames to be processed. Choosing a proper sampling rate is very important. A low sampling rate leads to poor video summaries; while a large sampling rate shortens the video summarization time. In VSCAN approach, the sampling rate used is selected to be one frame per second. So, for a video sample of duration one minute, and a frame rate of 30 fps (i.e., 1800 frames); the number of extracted frames is 60 frames.

\begin{figure}
\centering
\includegraphics[width=4.5in,height=2.5in]{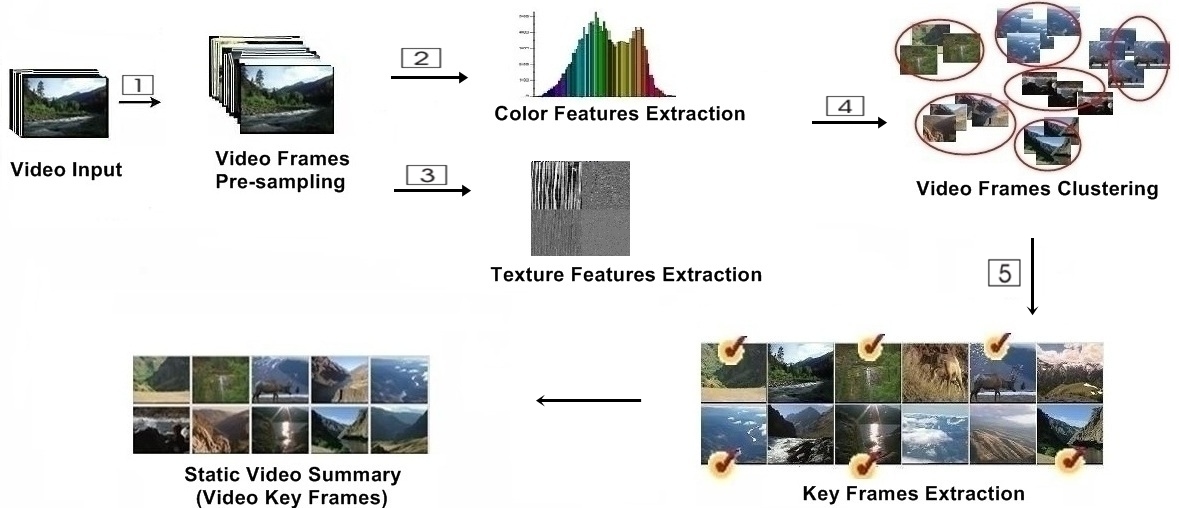}
\caption{VSCAN Approach}
\label{fig:image1}
\end{figure}

\subsection{Color Features Extraction}
In VSCAN, color histogram \cite{swain1991color} is applied to describe the visual content of video frames. In video summarization systems, the color space selected for histogram extraction should reflect the way in which humans perceive color. This can be achieved by using user-oriented color spaces as they employ the characteristics used by humans to distinguish one color from another \cite{stehling2002techniques,de2011vsumm}. One popular choice is the HSV color space, the HSV color space was developed to provide an intuitive representation of color and to be near to the way in which humans perceive color \cite{de2011vsumm}.

The color histogram used in VSCAN is computed from the HSV color space using 32 bins of H, 4 bins of S, and 2 bins of V. This quantization of the color histogram is established through experimental tests and aims at reducing the amount of data without losing important information.

\subsection{Texture Features Extraction}
Texture is a powerful low-level feature for representing images. It can be defined as an attribute representing the spatial arrangement of the pixels in a region or image \cite{IEEEStandardGlossary}.

Discrete Wavelet Transformation (DWT) is commonly used to extract texture features of an image by transforming it from spatial domain into frequency domain \cite{smith1994transform}. Wavelet transforms extract information from signal at different scales by passing the signal through low pass and high pass filters. Also, Wavelets provide multi-resolution capability and good energy compaction. In addition, they are robust with respect to color intensity shifts and can capture both texture and shape information efficiently \cite{singha2012signal}. 

In VSCAN, Discrete Haar Wavelet Transforms \cite{stankovic2003haar} is used to compute feature vectors as a texture representation for video frames, because it is fast to compute and also have been found to perform well in practice. Each video frame is divided into color channels and the Discrete Haar Wavelet Transform is applied on each channel. It is well known that the RGB color space is not suitable to reflect human perception of color 
\cite{de2011vsumm,girgensohn2001keyframe,liu2004shot}. Instead of using RGB, the video frame image is converted to HSV color space; moreover the video frame’s size is reduced into 64 X 64 pixels in order to reduce computation without losing significant image information. Next step is applying a two-dimensional Haar Wavelet transform on the reduced HSV image data with decomposition level 3. Finally the texture features of the video frames are extracted from the approximation coefficients of the Haar Wavelet Transforms.

\subsection{Video Frames Clustering}
DBSCAN (density-based spatial clustering of applications with noise) \cite{ester1996density} is a density-based algorithm which discovers clusters with arbitrary shape using minimal number of input parameters. The input parameters required for this algorithm is the radius of the cluster (Eps) and the minimum points required inside the cluster (Minpts) \cite{ester1996density}. 

Using DBSCAN clustering algorithm has many advantages. First, it does not require specifying the number of clusters in the data a priori, as opposed to partitioning algorithms like k-means \cite{parimala2011survey}. Second, DBSCAN can find arbitrarily shaped clusters. Third, it has a notion of noise. Finally, DBSCAN requires minimal number of input parameters.

In VSCAN, we apply a dual feature space DBSCAN algorithm. The proposed clustering algorithm used in VSCAN aims at adapting and modifying DBSCAN to be used by a video summarization system that utilizes both color and texture features. Instead of accepting only one input dataset as in the original DBSCAN, the clustering algorithm in VSCAN accepts both color and texture features of video frames as input datasets, with the Bhattacharya distance \cite{kailath1967divergence} as a dissimilarity measure. The Bhattacharyya distance between two discrete distributions P and Q of size n; is defined as:	 
\begin{equation}
 Bhattacharyya Distance = \sum\limits_{i=0}^{n}\sqrt{\sum{Pi}\bullet\sum{Qi}} 
\end{equation}
	
Selecting the Bhattacharyya distance as dissimilarity measure has many advantages \cite{aherne1998bhattacharyya}. First, the Bhattacharyya measure has a self-consistency property, as by using the Bhattacharyya measure all Poisson errors are forced to be constant therefore ensuring the minimum distance between two observations points is indeed a straight line \cite{aherne1998bhattacharyya}. The second advantage is the independence between Bhattacharyya measure and the histogram bin widths, as for the Bhattacharyya metric the contribution to the measure is the same irrespective of how the quantities are divided between bins; therefore it is unaffected by the distribution of data across the histogram \cite{aherne1998bhattacharyya}. Third advantage is that the Bhattacharyya measure is dimensionless; as it is not affected by the measurement scale used \cite{aherne1998bhattacharyya}.

The values of the Bhattacharyya distance between features vectors of two frames p and q, occurs between 0 and 1; where 0 means that two frames are completely not similar and 1 means that they are exact similar.

Original definitions of DBSCAN Algorithm can be found in \cite{ester1996density}. Following are the definitions of the proposed video clustering algorithm used by VSCAN .


\begin{definition} 
		\textbf{(CD- Color database)} a database containing color features extracted from video frames.
\end{definition}


\begin{definition} 
		\textbf{(TD- Texture database)} a database containing texture features extracted from video frames.
\end{definition}


\begin{definition} 	
    \textbf{(EpsColor-color-based similarity of video frame)}\\EpsColor-color-based similarity of video frame p, denoted by $S_{EpsColor}(p) $ is defined by: 	
		$	
				S_{EpsColor}(p) = \left\{q \in CD\vert BhatDist(p,q) \geq EpsColor \right\},
		$
		where BhatDist is Bhattacharyya distance.
\end{definition}


\begin{definition} 	
		\textbf{(EpsTexture-texture-based similarity of video frame)}\\EpsTexture-texture-based similarity of video frame p, denoted by $S_{EpsTexture}(p)$ is defined by:		
		$		
				S_{EpsTexture}(p) = \left\{q \in TD\vert BhatDist(p,q) \geq  EpsTexture \right\},				
		$
		where BhatDist is Bhattacharyya distance.
\end{definition}


\begin{definition} 	
 \textbf{(Eps composite similarity score of a video frame)}\\ Eps composite similarity of a video frame p, denoted by $S_{Eps}(p)$ is defined by:		
		$
			S_{Eps}(p) = \lbrace q \in CD,TD\vert score(p,q ) = Eps \rbrace, where \ Eps \ \& \ score(p,q) \in \lbrace 0, 1, 2 \rbrace
		$
		in which possible values are defined as follows:			
		\begin{description}
			\item $ \textbf{0}:if \ p \notin S_{EpsColor}(q) \ AND \ p \notin S_{EpsTexture}(q),$ in this case p,q are NOT similar.

			\item $ \textbf{1}:if \ p \in S_{EpsColor}(q) \ OR \ p \in S_{EpsTexture}(q),$ in this case p,q are color-based similar OR texture-based similar.				
			
			\item $ \textbf{2}:if \ p \in S_{EpsColor}(q) \ AND \ p \in S_{EpsTexture}(q),$	in this case p,q are color-based similar AND texture-based similar.			
		\end{description}		
		
\end{definition}

\begin{definition} 	
\textbf{(Directly-similar)} A frame p is directly-similar to a frame q wrt. Eps, MinPts if 
	$	p  \in S_{Eps}(q)\emph{ and } \vert S_{Eps}(q) \vert \geq MinPts$ (\textbf{core frame condition}).
\end{definition}

\begin{definition} 
\textbf{(Indirectly-similar)}  A frame p is indirectly-similar to a frame q wrt. Eps and MinPts if there is a chain of frames
	$ p_{1},...., p_{n}, \ p_{1} = q, \ p_{n} = p \ such \ that \ p_{i+1} \ is \ directly \ similar \ to \ p_{i}$
\end{definition}

\begin{definition} 
\textbf{(Connected-similar)} A frame p is connected-similar to a frame q wrt. Eps and MinPts, if there is a frame o such that both, p and q are indirectly-similar to o wrt.Eps and MinPts.
\end{definition}

\begin{definition}
\textbf{(Video cluster)} Let D be a database of frames. A cluster C wrt. Eps and MinPts is a non-empty subset of D satisfying the following conditions:

\begin{itemize}
	\item 	$ \forall \ p,q: \ if \ p \in C$ and q is indirectly-similar to p wrt. Eps and MinPts, then q $\in C $ \textbf{(Maximality)}.
	\item   $ \forall \ p,q \in C:$ p is connected-similar to q wrt. Eps and MinPts \textbf{(Connectivity)}.
\end{itemize}

\end{definition}

\begin{definition}
\textbf{(Noise)} Let 
$ C_{1},...,C_{k} $ \emph{be the video clusters of the database D. The noise is defined as the set of frames in the database D not belonging to any cluster} $ C_{i}$, \emph{i=1,..,k. i.e. Noise = } $ \lbrace p \in D \vert \forall i : p \notin C_{i} \rbrace $ 
\end{definition}

The steps involved in VSCAN clustering algorithm are as follows:

\begin{enumerate}
	\item Select an arbitrary frame p.
	\item Retrieve all frames that are indirectly-similar from p w.r.t. Eps and Minpts.
	\item If p is a core frame, a video cluster is formed. 
	\item If p is a border frame, no frames are indirectly-similar from p and the next frame of the database is visited.
	\item Continue the process until all the frames have been processed. 
\end{enumerate}

For clustering the video frames, we apply the proposed clustering Algorithm on extracted color and texture features of the pre-sampled video frames. According to our experimental tests, we set the input parameters in VSCAN algorithm as follows: EpsColor = 0.97, EpsTexture = 0.97, Eps = 2 and Minpts = 1. As per provided definitions, the EpsColor is the Bhattacharyya distance threshold for grouping frames using color features, this means that only frames with Bhattacharyya distance greater than or equal to 0.97 are color-based similar and  eligible to be in the same cluster. While EpsTexture is the Bhattacharyya distance threshold for grouping frames using texture features, this means that only frames with Bhattacharyya distance greater than or equal to 0.97 are texture-based similar and eligible to be in the same cluster.

As per the clustering algorithm definitions, setting Eps value to 2 means that video frames are eligible to belong in same cluster if they are color-based similar and texture-based similar. While Minpts input parameter value is the key value for noise elimination mechanism, Minpts in the algorithm is the minimum number of neighbor frames allowed to create a cluster within the current frame, i.e. setting Minpts to 1, means that minimum cluster size equals to 2 and any cluster of size 1 will be considered as a noise. Since we have selected sampling rate of 1 frame per second, setting Minpts to 1 is equivalent to discarding those video segments of duration less than 2 seconds.

\subsection{Key Frames Extraction}
After clustering the video frames, the final step is selecting the key frames from the video clusters. In this step the noise frames are discarded and then for each cluster the middle core frame in the ordered frames sequence is selected to construct the video summary. According to our experiments we found that this middle core frame usually is the best representative of the cluster to which it belongs.

\section{Experimental Evaluation}
In this paper, a modified version of an evaluation method Comparison of User Summaries (CUS) described in \cite{de2011vsumm} is used to evaluate the quality of video summaries. In CUS method, the video summary is built manually by a number of users from the sampled frames and the user summaries are taken as reference (i.e. ground truth) to be compared with the automatic summaries obtained by different methods \cite{de2011vsumm}.

The modifications proposed to CUS method aims at providing a more perceptual assessment of the quality of the automatic video summaries. Instead of comparing frames from different summaries using color features only as in CUS method, both color and texture features (as in section 3.2 and section 3.3) are used to detect the similarity of the frames. Once two frames are color-based similar or texture-based similar, they are excluded from the next iteration of the comparison process. In this modified CUS version, the Bhattacharya distance is used to detect both color and texture similarity; in this case the distance threshold value for color and texture similarity is set to 0.97.

In order to evaluate the automatic video summary, the F-measure is used as a metric. The F-measure consolidates both Precision and Recall values into one value using the harmonic mean \cite{blanken2007multimedia}, and it is defined as:

\begin{equation}
F \textrm{-} measure = \frac{2 \times Precision \times Recall}{Precision + Recall} 
\end{equation}

The Precision measure of video summary is defined as the ratio of the total number of color-based similar frames and texture-based similar frames to the total number of frames in the automatic summary; and the Recall measure is defined as the ratio of the total number of color-based similar frames and texture-based similar frames to the total number of frames in the user summary

VSCAN approach is evaluated on a set of 50 videos selected from the Open Video Project  \footnote{Open Video Project. \url{http://www.open-video.org}}. All videos are in MPEG-1 format (30 fps, 352 × 240 pixels). They are distributed among several genres (documentary, historical, lecture, educational) and their duration varies from 1 to 4 min. Also, we use the same user summaries used in \cite{de2011vsumm} as a ground-truth data. These user summaries were created by 50 users, each one dealing with 5 videos, meaning that each video has 5 summaries created by five different users. So, the total number of video summaries created by the users is 250 summaries and each user may create different summary. 

For comparing VSCAN approach with other approaches, we used the results reported by three approaches: VSUMM \cite{de2011vsumm}, STIMO \cite{furini2010stimo}, and DT \cite{mundur2006keyframe}. In addition to that, the automatic video summaries generated by our approach were compared with the OV summaries generated by the algorithm in \cite{dementhon1998video}. All the videos, user summaries, and automatic summaries are available publicly 
\footnote{\url{http://sites.google.com/site/vscansite/}}.

In addition to previous approaches, we implemented a video summarization approach called DB-Color using the original DBSCAN algorithm with the color features only as an input. We used the same color features extraction method used in the proposed VSCAN approach as in section 3.2 and also the same input parameters for color similarity and noise detection as in section 3.3, i.e. Eps = 0.97 and Minpts = 1. The reason for implementing DB-Color is to test the effect of using color only instead of combining both color and texture as in VSCAN.

\tablename{ 1} shows the mean F-measure achieved by the different video summarization approaches. The results indicate that VSCAN performs better than all other approaches. Also, we notice that combining both color and texture features together as done in VSCAN gives better results than using color features only as in DB-Color. However, DB-Color achieved better results if compared to the other four approaches (OV, DT, STIMO, and VSUMM), which indicates that using DBSCAN clustering algorithm is efficient for generating static video summary.

\begin{table}[!t]
\centering
\caption{Mean F-measure achieved by different approaches}
\begin{tabular}{l l l l l l l} 
\hline\noalign{\smallskip}
\textbf{Approach} \ \ & OV \ \ \ & DT \ \ \ & STIMO \ \ & VSUMM \ \ & VSCAN \ \ & DB-Color \ \ \\
\noalign{\smallskip}
\hline
\noalign{\smallskip}
\textbf{Mean F-Measure} \ \ & 0.67 & 0.61 & 0.65 & 0.72 & \textbf{0.77} & 0.74 \\
\hline
\end{tabular}
\end{table}

\section{Conclusion}
In this paper, we presented VSCAN, a novel approach for generating static video summaries. VSCAN utilizes a modified DBSCAN clustering algorithm to summarize the video content using both color and texture features of the video frames. Combining both color and texture features enabled VSCAN to overcome the drawback of using color features only as in other approaches. Also, as an advantage of using a density-based clustering algorithm, VSCAN could overcome the drawback of determining a priori number of clusters; thus, the extra step needed for estimating the number of clusters is avoided. Also, as an advantage of using a modified DBSCAN algorithm, VSCAN can detect noise frames automatically without extra computations.

As an additional contribution, we proposed an enhanced evaluation method based on color and texture matching. The main advantage of this evaluation method is to provide a more perceptual assessment of the quality of automatic video summaries.

Future work includes combining other features to VSCAN approach like edge and motion descriptors. Also, another interesting future work could be generating video skims (dynamic key frames, e.g.\ movie trailers) from the extracted key frames. Since the video summarization step is usually considered as a prerequisite for video skimming \cite{truong2007video}, the extracted key frames from VSCAN can be used to develop an enhanced video skimming system.


\bibliographystyle{splncs} 
\bibliography{VSCANBib}

\end{document}